\documentclass[letterpaper, 10 pt, conference]{ieeeconf}  
\IEEEoverridecommandlockouts                              
\usepackage{graphicx} 
\usepackage{subfigure}	
\usepackage[table]{xcolor}
\usepackage{diagbox}
\usepackage{multirow}
\usepackage{amsmath}
\usepackage{amsfonts,amssymb}
\usepackage{color}
\usepackage{cite}
\usepackage{booktabs}
\usepackage{pifont}
\usepackage[capitalize]{cleveref}
\usepackage{subcaption}

\title{\LARGE \bf
FBPT: A Fully Binary Point Transformer 
}

\author{Zhixing Hou$^{1}$, Yuzhang Shang$^{2}$, Yan Yan$^{2\dag}$ 
\thanks{$^{\dag}$ Corresponding author.}
\thanks{$^{1}$ Zhixing Hou is with the School of Computer Science and Engineering, Nanjing University of Science and Technology, Nanjing, China. {\tt\small E-mail: hzx@njust.edu.cn}} 
\thanks{$^{2}$ Yuzhang Shang and Yan Yan are with the Department of Computer Science, Illinois Institute of Technology, Chicago, USA. {\tt\small E-mail:  yshang4@hawk.iit.edu, yyan34@iit.edu} (*Corresponding author)}
}

\begin{document}

\maketitle

\begin{abstract}
     This paper presents a novel Fully Binary Point Cloud Transformer (FBPT) model which has the potential to be widely applied and expanded in the fields of robotics and mobile devices. By compressing the weights and activations of a 32-bit full-precision network to 1-bit binary values, the proposed binary point cloud Transformer network significantly reduces the storage footprint and computational resource requirements of neural network models for point cloud processing tasks, compared to full-precision point cloud networks. However, achieving a fully binary point cloud Transformer network, where all parts except the modules specific to the task are binary, poses challenges and bottlenecks in quantizing the activations of Q, K, V and self-attention in the attention module, as they do not adhere to simple probability distributions and can vary with input data. Furthermore, in our network, the binary attention module undergoes a degradation of the self-attention module due to the uniform distribution that occurs after the softmax operation.  The primary focus of this paper is on addressing the performance degradation issue caused by the use of binary point cloud Transformer modules. We propose a novel binarization mechanism called dynamic-static hybridization. Specifically, our approach combines static binarization of the overall network model with fine granularity dynamic binarization of data-sensitive components. Furthermore, we make use of a novel hierarchical training scheme to obtain the optimal model and binarization parameters. These above improvements allow the proposed binarization method to outperform binarization methods applied to convolution neural networks when used in point cloud Transformer structures.
     To demonstrate the superiority of our algorithm, we conducted experiments on two different tasks: point cloud classification and place recognition. In point cloud classification,  our model achieved an accuracy of 90.9\%, which is only a 2.3\% decrease compared to the full precision network. For the place recognition task, we achieved 91.02\% in the top @1\% metric and 82.87\% in the top @1 metric on the Oxford RobotCar dataset in terms of the average recall rate. Moreover, our model exhibits a significant reduction of over 80\% in terms of model size and FLOPs (floating-point operations) compared to the baseline.
    \end{abstract}
    
    %===============================================================================
    
    \section{Introduction}

    In recent years, there have been significant advancements in the research fields of robot perception and artificial intelligence, and one of the key contributing factors is the discovery of the transformer~\cite{vaswani2017attention} structure. The transformer model has greatly enhanced the performance of various tasks such as natural language processing~\cite{devlin2019bert, brown2020language}, image recognition~\cite{dosovitskiy2020vit, liu2021swin}, and point cloud processing~\cite{zhao2021point, guo2021pct}, particularly in large-scale pre-trained tasks. However, a major limitation of applying these transformer models in real-world scenarios is their high memory and computational requirements, which hinders their widespread deployment in mobile or wearable device contexts.

    To overcome these limitations, Transformer models used in various tasks need to be compressed while still maintaining good performance. Previous studies have proposed various Transformer compression methods, including pruning~\cite{fan2019reducing,michel2019sixteen,voita2019analyzing}, knowledge distillation~\cite{jiao2020tinybert,sun2019patient,sanh2019distilbert}, low-rank approximation~\cite{ma2019tensorized,lan2019albert}, and adaptive dynamic networks~\cite{hou2020dynabert,liu2020fastbert}. The above methods reduce the network size and increase efficiency, while some papers focus on quantization~\cite{zafrir2019q8bert,zhang2020ternarybert,shen2020q,lizhexin2022qvit,liyanjing2022qvit,yuan2023rptq,shang2023post} as an attractive and alternative method that has achieved great success in practice. Among the quantization methods, extreme quantization~\cite{bai2021binarybert, qin2021bibert, liu2022bit} pushes the limits of transformer compression by converting the original 32-bit full-precision values into 1-bit binarized values. In binary quantization methods, the model size is substantially reduced, and the calculation speed is also accelerated using binary bitwise operations~\cite{rastegari2016xnor} such as $xnor$ and $bitcount$. This provides a significant advantage over non-binary quantization methods.

    \begin{figure}[t]
      \centering
       \includegraphics[width=0.5\linewidth]{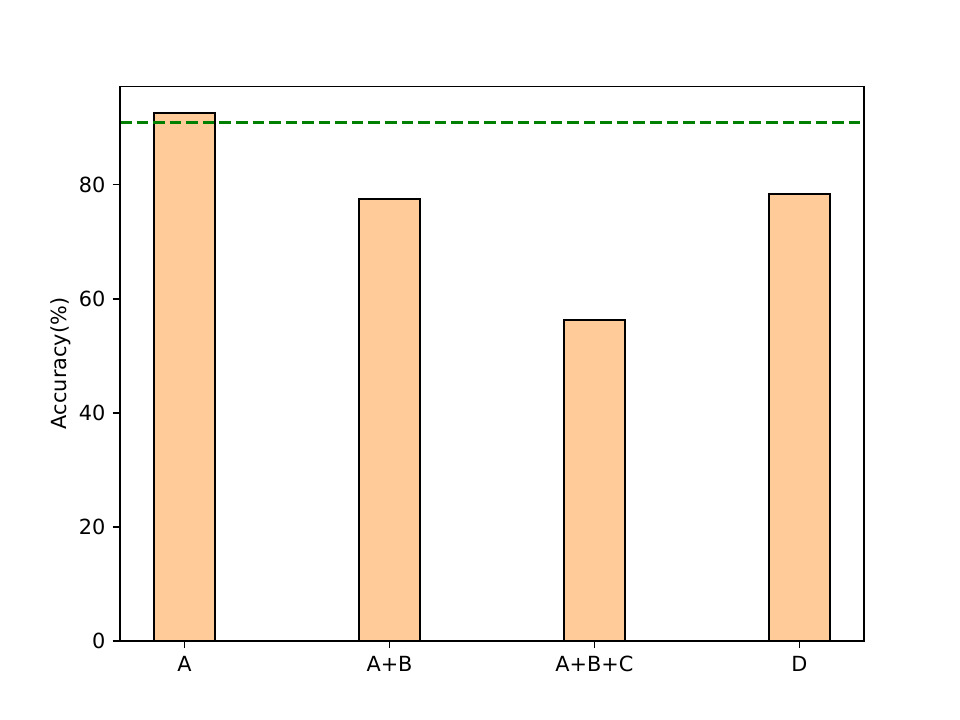}
       \caption{degradation analysis (without local feature skip concatenation). $A$: binary local feature, $B$: binary weights in transformer module, $C$:binary activations in transformer module, and $D$: binary transformer module. Green dashed lines indicate the results of the full-precision PCT network.}
       \label{fig:degradation}
    \end{figure}

    In addition to natural language and image processing, the analysis and processing of 3D point clouds are also crucial in artificial intelligence research. Unlike natural language and images, 3D point clouds possess unique characteristics such as irregular formats, uneven density, and coordinate system dependency. As a result, the point cloud transformer model \cite{zhao2021point, guo2021pct, wu2022point} differs from the classic transformer models used in NLP and computer vision domains. This distinction will be described in~\ref{subsec:fp-pct}. Given these differences, the binarization of point cloud transformers necessitates a redesigned solution. 

    To our knowledge, no \textbf{fully binarized point cloud transformer} has been proposed thus far. As the \textbf{first work} to implement this research, we conducted binarization on different parts of the network to identify the modules most sensitive to binarization demonstrated in \cref{fig:degradation}. In order to avoid significant performance degradation, we propose a novel binarization mechanism that applies various granularities of binarization to different modules in a dynamic manner. 
    Finally, based on the full-precision point cloud transformer network~\cite{guo2021pct}, we develop a fully binarized network replica that includes both weight and activation binarization.

    However, in the fully binarized point cloud transformer network, the self-attention mechanism degenerates to uniformly distributed after softmax, which leads to a loss of weighted attention towards the $Value$ in Transformer. To prevent this degradation, we design a novel hierarchical training scheme to gradually determine model parameters and binarization parameters in three stages.

    Moreover, we find that in the full-precision point cloud transformer network, vanilla self-attention has limited impact on network accuracy. However, after incorporating binarized self-attention, significant improvements in accuracy are observed, along with a certain ability to suppress overfitting.

    We impose the proposed binary point cloud transformer model to the point cloud classification task and evaluate its performance on the ModelNet40 dataset~\cite{wu20153d}. The results show that our model achieve an overall accuracy of 90.9\%, with only a 2.3\% performance drop compared to the full-precision counterpart, while reducing the model size by 87.2\% and the FLOPs (floating-point operations) by 80.2\%.
    To further validate the superiority of our proposed binary point cloud transformer, we integrate it with metric learning and impose it on the place recognition task. Our model achieves an accuracy of 91.02\% in the \textit{top1\%} metric and 82.87\% in the \textit{top1} metric on the Oxford RobotCar dataset~\cite{RobotCarDatasetIJRR}.

    In summary, the contributions and novelties of this paper can be summarized as follows:
    \begin{itemize}
        \item We introduce a novel, Fully Binary Point Cloud Transformer model called FBPT.
        \item Through experimental analysis, we identify performance bottlenecks in binarized point cloud Transformers and propose novel techniques to address these bottlenecks, including fine-grained binarization, dynamic-static hybridization binarization, and a hierarchical training scheme.
        \item We demonstrate the efficacy of our proposed FBPT model through experiments on point cloud classification and place recognition tasks. Our results reveal that the FBPT model exhibits excellent performance in terms of accuracy, model size, computational complexity, and generalization capabilities.
    \end{itemize}

%%%%%%%%%%%%%%%%%%%%%%%%%%%%%%%%%%%%%%%%%%%%%%%%%%%%%%%%%%%%%%%%%%%
	
\section{Related works}
\label{sec:related_works}
    \subsection{Model Binarization}
    \label{subsec:binary_quant}
    Courbariaux et al.\cite{courbariaux2016binarized} first proposed the concept of the binary neural network, through two kinds of state quantities +1 and -1 to fit full-precision model parameters and activation values, achieved the effect of compressing the model and improving the operation efficiency. Subsequently, Rastegari et al. \cite{rastegari2016xnor} introduced XNOR-Net, and demonstrated the effectiveness of the model through rigorous theoretical analysis and experiments on large classification datasets such as ImageNet1K. This approach can theoretically reduce the model size by $1/32$ and speed up inference by 58 times.

    In the research on Transformer binarization,
    first, Bai et al. \cite{bai2021binarybert} proposed a transformer binarization work named BinaryBERT by equivalently splitting from a half-size ternary BERT network \cite{zhang2020ternarybert} that pushes the transformer quantization to the limit. Qin et al. \cite{qin2021bibert} proposed BiBERT which introduces the Bi-Attention module and Direction-Matching distillation method from the perspective of information theory by maximizing the information entropy of the binarized representations. Liu et al. \cite{liu2022bit} also implemented a binarized BERT model called BiT that includes a two-set binarization scheme that utilizes different binarization methods to compact activations with and without non-linear activation layers. Furthermore, they made use of the multi-step distillation method, which first distills the full-precision model into an intermediate model and then distills the intermediate model into the binarized transformer model.
    
    As for vision transformer research fields, Li et al. \cite{lizhexin2022qvit} compacted parameters and activations of the vision transformer model to 3 bit and Li et al. \cite{liyanjing2022qvit} further pushed the limit of vision transformer quantization to 2-bit. However, there has been no binarized vision transformer, i.e. 1-bit weights and activation representations proposed so far. 
    
    In the same situation as the vision transformer, few binarization works are applied to point cloud processing tasks. The first binary neural network for raw point cloud processing proposed by Qin et al. \cite{qin2020bipointnet} introduced Entropy-Maximizing Aggregation and Layer-wise Scale Recovery to effectively prevent immense performance drop. Xu et al. \cite{xu2021poem} introduced a bi-modal distribution concept to mitigate the impact of small disturbances and exploited Expectation-Maximization to constrain the weights of the network into this distribution. Su et al. \cite{su2022svnet} designed a binarization and training method for the point cloud network with invariant scales and equivariant vectors simultaneously. But these three works are all based on the MLP structure originated from the classic work PointNet \cite{pointnet} without the participation of the transformer structure. So our work proposed in this paper is the first binary point cloud transformer model and it is also the first time to be applied to place recognition.

    \subsection{Dynamic and Static Quantization}
    \label{subsec:dsq}
    In the research field of model quantization, post-training quantization refers to the process of quantizing the floating-point weights and activations in a trained model into fixed-point integers with a specified number of bits. This is achieved by designing quantizers and utilizing a small amount of calibration data to compute quantization parameters. These parameters are then used to convert the model's weights and activations into their corresponding fixed-point representations. Depending on the different methods used to calculate quantization parameters for activations, the quantization process can be categorized into dynamic quantization and static quantization. In static quantization, quantization parameters of activations are determined offline using the activations of calibration samples. This approach allows for precise control over the quantization process based on a predefined set of activation values. On the other hand, dynamic quantization utilizes runtime statistics of activations to calculate quantization parameters. This method adapts the quantization parameters during runtime, taking into account the actual distribution of activations encountered during inference.

    \subsection{Quantization Granularity}
    \label{subsec:qg}
    Quantization can be done at different granularity levels. Per-tensor quantization applies a single step size to the entire matrix. For finer-grained quantization, per-token quantization assigns different step sizes to activations associated with each token, while per-channel quantization assigns different step sizes to each output channel of weights. Another option is group-wise quantization, which uses different step sizes for different channel groups.

    \begin{figure}[t]
        \centering
        \subfigure[]{\label{fig:pct_a}\includegraphics[width=0.16\textwidth]{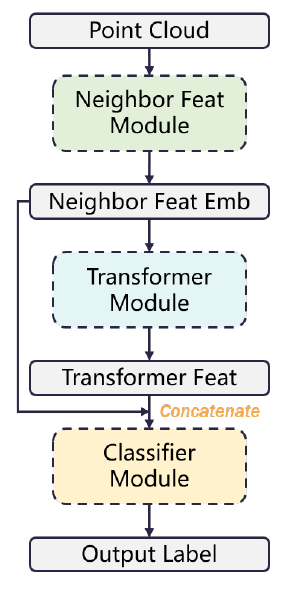}}
        \subfigure[]{\label{fig:pct_b}\includegraphics[width=0.16\textwidth]{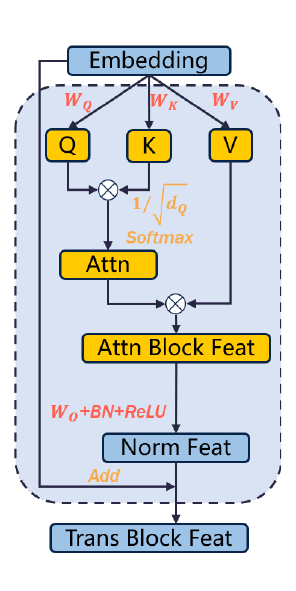}}
        \caption{(a) Point cloud transformer model used for classification. (b) Full-precision point cloud transformer block, which is the major structure in Transformer module.}
	\label{fig:pct}
    \end{figure}

    \subsection{Full-precision Point Cloud Transformer}
    \label{subsec:fp-pct}
    There are several different types of transformer-related works to deal with point cloud tasks. Among them, the most typical works are Point Transformer \cite{zhao2021point}, Point Transformer V2 \cite{wu2022point}, and PCT (Point Cloud Transformer) \cite{guo2021pct}. Due to its clear, concise, and effective structure shown in \cref{fig:pct}, the PCT model \cite{guo2021pct} is chosen as the full-precision model counterpart of our binary model. We divide the whole pipeline into three main components and name them the neighbor feature extraction module, the transformer module, and the classifier module.

    \begin{figure}[t]
        \centering
        \subfigure[]{\label{fig:bit_op}\includegraphics[width=0.18\textwidth]{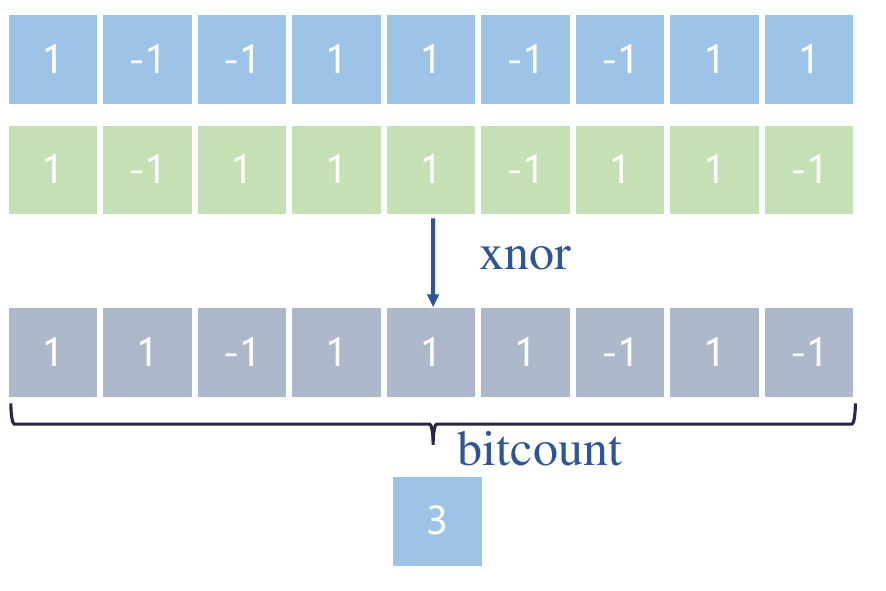}}
        \subfigure[]{\label{fig:pct_e}\includegraphics[width=0.2\textwidth]{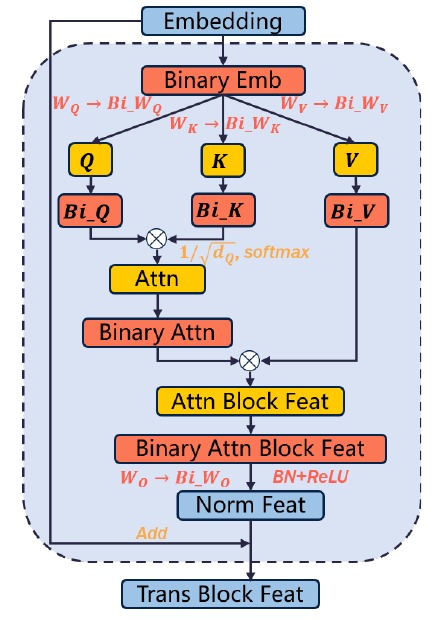}}
        \caption{(a) Bitwise binary operations: $xnor$ and $bitcount$. (b) Binary point cloud transformer block.  The symbol $\otimes$ in binary transformer blocks indicates binary operation including $xnor$ and $bitcount$, equivalent to dot product operation in full-precision transformer blocks.}
        \label{fig:transformer_model}
    \end{figure}

	%===============================================================================
\section{Fully binary point cloud transformer}
\label{sec:fbpt}

    \begin{figure*}[t]
      \centering
       \includegraphics[width=0.92\textwidth]{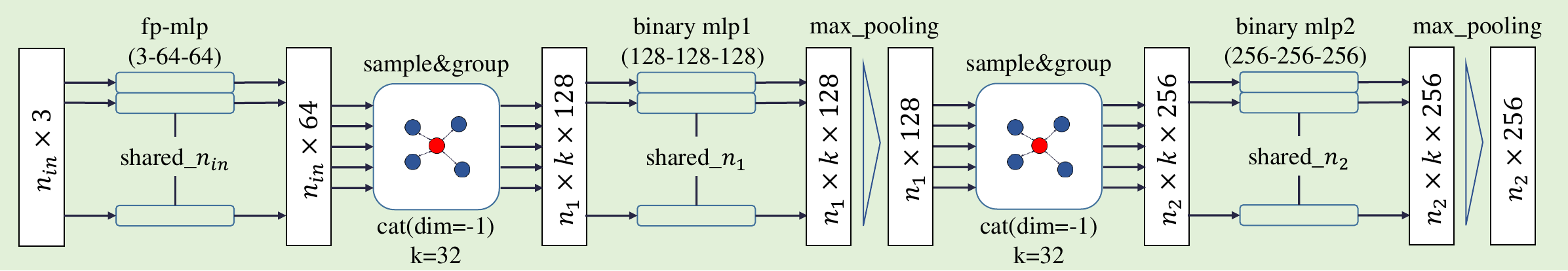} 
       \caption{The local feature module. The input point will be preserved as full precision and transformed to point-wise features with full precision MLP, while the point embedding will be fed into the transformer module generated with binary MLP following sampling and grouping.}
       \label{fig:neighbor_module}
    \end{figure*}
    
    \subsection{Revisit Binarization for Linear Layers}
    \label{subsec:bi_linear}
    Linear transformation is a fundamental operation in transformer networks. In this section, we will focus on introducing binarization techniques that are specifically designed for linear layers. The definition of linear layers and their binarization in neural networks is as follows:
    \begin{equation}
        \begin{aligned}
        \mathbf{Y} &= \mathbf{W_r} \ast \mathbf{A_r} \\
        &= \alpha \mathcal{B}(\mathbf{W_r^{'}}) \ast \beta \mathcal{B}(\mathbf{A_r^{'}}) \\
        &= \alpha \beta (\mathcal{B}(\mathbf{W_r^{'}}) \otimes (\mathcal{B}(\mathbf{A_r^{'}})))\\
        &= \alpha\beta \mathbf{W_b} \otimes \mathbf{A_b}, \\
        \end{aligned}
        \label{eq:bi_linear}
    \end{equation}
    where the output of the linear layer is denoted as the matrix $\mathbf{Y}$, obtained by multiplying the weight matrix $\mathbf{W_r}$ with the input activation matrix $\mathbf{A_r}$. Here, $\mathbf{W_r}$ represents the weight of the linear layer, and $\mathbf{A_r}$ refers to the activation of the linear layer. The subscript $\mathbf{r}$ indicates that these matrices have real-valued floating-point precision. Before binarization, $\mathbf{W_r}$ and $\mathbf{A_r}$ are transformed into  $\mathbf{W_r^{'}}$ and $\mathbf{A_r^{'}}$ respectively through a shift and scale operation. The binarization function $\mathcal{B(\cdot)}$ is applied to $\mathbf{W_r^{'}}$ and $\mathbf{A_r^{'}}$ based on the specific format of the input. Once the weights and activations are binarized, the original matrix multiplication (denoted as $\ast$) is replaced by binary bitwise operations (represented by $\otimes$). These operations include the $xnor$ operation and the $bitcount$ operation, as shown in Fig.~\ref{fig:bit_op}.

    The scale factors $\alpha$ and $\beta$ can be extracted from matrix multiplication and precomputed as a global coefficient for the overall result of binarized matrix operations, following the XNOR-Net~\cite{rastegari2016xnor}. The calculation of these scale factors is described as follows:
    \begin{equation}
        \begin{aligned}
        \alpha\beta&=\frac{\sum{|W_i||A_i|}}{n}\\
        &\approx (\frac{1}{n}||W||_{l1})(\frac{1}{n}||A||_{l1}),
        \end{aligned}
        \label{eq:scale_factor}
    \end{equation}
    where the subscript $l1$ denotes the $l1$-$norm$. To better fit real values with binarized values, various binarization schemes can be employed based on the format of activations, as described in the BiT paper \cite{liu2022bit}, depicted in Eq.~\ref{eq:bi_sign} and \ref{eq:bi_round}. For layers where activations are distributed from negative to positive values, the $sign$ function is used for binarization. On the other hand, for layers with activations confined to the range of positive values only (e.g., after $softmax$ function), the $round$ function is used. In such cases, the real values are binarized into the set $\{0,1\}$:
    \begin{equation}
      \mathbf{A_b} = \left\{\begin{array}{ccl}
            1 & \mbox{for}
            & x\geq 0 \\ -1 & \mbox{for} & otherwise \\
                \end{array} \right.
      \label{eq:bi_sign}
    \end{equation}
    and for the layers only distributed on the range of positive values:
    \begin{equation}
      \mathbf{A_b} = \left\{\begin{array}{ccl}
            1 & \mbox{for}
            & x\geq 0.5 \\ 0 & \mbox{for} & otherwise \\
                \end{array} \right.
      \label{eq:bi_round}
    \end{equation}

    \subsection{Local Feature Module and Binarization}
    \label{subsec:bi_local}
    The local feature extraction module, illustrated in Fig.~\ref{fig:neighbor_module}, is comprised of a series of blocks that are interconnected. Each block consists of several components, including shared multi-layer perceptron (MLP) networks, sampling and grouping operations, and pooling layers. The fundamental idea of this local features method can be attributed to the DGCNN~\cite{Wang2019DynamicGC}.
    
    Following conventional practices, certain elements in the module are retained as full-precision representations. Specifically, the input point cloud data and the MLP in the initial layer remain in their original form, without any quantization. However, for the remaining components within the module, such as activations and weights, a binarization process is applied resulting in a binary MLP representation. By employing this approach, we can effectively extract local features from the data while achieving a fully binarized module. 

    \subsection{Bottleneck of FBPT}
    \label{subsec:bottleneck}
    We independently binarized the corresponding parts in the network shown in Fig~\ref{fig:degradation}. we discovered that binarizing the transformer module leads to a significant decline in model performance. We can summarize the reasons for this performance drop into three aspects.

        \begin{itemize}
        \item Firstly, the self-attention part in the transformer essentially weights the values, and larger weights have a greater impact on the results. However, after passing through the softmax function, most of the values in the self-attention tend to cluster around zero, with fewer significant outliers. Conventional binarization methods would eliminate the influence of these outliers on the model.
        \item Secondly, the linear operations in the transformer not only involve computations between weights and activations but also computations between different activations, such as matrix multiplication between $Query$ and $Key$. Additionally, the distribution of activation values in the transformer varies significantly with many outliers, which change with the input. Therefore, fitting fixed binarization parameters to accurately represent the full precision activations is challenging.
        \item Finally, in the proposed fully binarized point cloud transformer network, the self-attention degenerates to uniformly distributed after the $softmax$ function, which leads to a loss of weighted attention towards the $Value$ in Transformer.
    \end{itemize}

    To address these issues, we proposed the following three improvements: fine-grained binarization (\cref{subsec:fine-granularity}), dynamic-static hybridization binarization (\cref{subsec:d_s_h_b}), and hierarchical training scheme (\cref{subsec:hierarchical}).

    \subsection{Fine-grained Binarization for Self-attention}
    \label{subsec:fine-granularity}

        We have increased the granularity of quantization for activations in the self-attention block. We believe that simply binarizing these activations, which are highly sensitive to data and exhibit complex distributions (such as query, key, value, and attention), would severely impact model performance. However, to effectively leverage the advantages of binary operations, we apply a higher granularity of binarization to these activations. Specifically, we first determine the range of activation values and then select appropriate partition points uniformly or non-uniformly across the entire range. This forms multiple groups of masks on the activation tensor, where each group shares the same set of binarization parameters. By using finer-grained binarization, the quantization bit-width of activations across the entire tensor is expanded from 1 bit to 2 bits or even 3 bits, while internally consisting of individual binary values, thereby ensuring smooth execution of binary operations.
    \subsection{Dynamic-static Hybridization Binarization}
    \label{subsec:d_s_h_b}

        The proposed binarization method is inspired by the concepts of dynamic and static post-training quantization, which is introduce in \ref{subsec:dsq}. We define static binarization as the utilization of binary activations during the training process, with optimal binarization parameters being learned through back propagation. In contrast, dynamic binarization entails retaining real-value activations during training while dynamically computing the parameters required for activation binarization during inference.

        Specifically, for activations that are highly sensitive to inputs, such as $Query$, $Key$, and $Value$, we no longer used pre-trained binarization parameters. Instead, we calculated the binarization parameters in real-time during algorithm execution. This approach allows the binarization parameters to be dynamically adjusted according to changes in the input. Although calculating binarization parameters dynamically increases computational overhead, the proportion of activations requiring dynamic computation is relatively small. Binarized activations can be operated with binarized weights, thereby improving the execution speed of the algorithm.
    
    \subsection{Hierarchical Training Scheme}
    \label{subsec:hierarchical}
        Training the FBPT model directly, similar to other network models, may result in self-attention being suppressed and becoming excessively small due to the presence of residual connections. Consequently, after passing through the softmax function, the self-attention weights degrade into a uniform distribution, rendering the weighting capacity of self-attention ineffective. To address this issue, a hierarchical training scheme is proposed which divides the training into three stages: non-transformer modules including local feature extraction module and parts of linear layer in the Transformer module, weights of the transformer module, and activations of the transformer module. 
        Since self-attention is a crucial structure in the transformer, we perform static binarization on non-transformer modules in the first stage while freezing network parameters and binarization parameters.
        In the second stage, we apply static binarization to the weights of the transformer module. Subsequently, in the third stage, we employ dynamic binarization for the activations of the transformer module. The first two stages are completed during the training process, while the third stage involves real-time computation during inference. Due to the effect of these three stages on the determination of binarization parameters, we collectively refer to them as the hierarchical training scheme.

%==============================================================================
\section{Experiments}
\label{sec:result}

    \subsection{Experimental Settings}
    \label{subsec:exp_settings}
    
    \noindent\textbf{Classification dataset}: ModelNet40~\cite{wu20153d} dataset. This dataset is the most popular benchmark for point cloud classification. A total of 12,311 CAD models from the 40 categories are split into 9,843 for training and 2,468 for testing.
    
    \noindent\textbf{Place recognition datasets:} We use the benchmark datasets created in \cite{Uy2018PointNetVLADDP} to train the FBPT model for the place recognition task. This frequently-used benchmark contains a modified Oxford RobotCar dataset \cite{RobotCarDatasetIJRR} for training and testing, as well as three smaller in-house datasets: University Sector (U.S.), Residential Area (R.A.), and Business District (B.D.) only for testing. The dataset settings are also the same as the ones in \cite{Uy2018PointNetVLADDP}.

    \noindent\textbf{Evaluation metrics:} The same as the previous works, overall accuracy (OA) and mean accuracy (mAcc) are used to evaluate the classification performance of the proposed model. While the top 1\% (@1\%) and top 1 (@1) average recall rate as the evaluation metrics are adopted to evaluate the place recognition performance of the algorithm.
 
    \noindent\textbf{Classification implementation details:}  
    The proposed FBPT model is trained from scratch without leveraging any pre-trained model. We use the stochastic gradient descent (SGD) optimizer and an initial learning rate of 0.01 with 0.9 momentum to train our network $400+100$ epochs (two stages) iteratively on the Pytorch platform. A cosine annealing schedule is adopted to adjust the learning rate. The mini-batch size is set to 32. All the evaluations are conducted on an NVIDIA RTX2080Ti GPU card.
    
	\begin{table}[t]
    	\caption{Comparison results for point cloud classification.} 
		\label{tab:result_tmp}
		\centering
            \rowcolors{2}{gray!50}{gray!20}
		\begin{tabular}{c|c|c|c|c}  %确定表格竖行格式   设置列宽度：将c换为p{ cm}
			\hline       
			\hline
			\rowcolor{red!30}Methods         &Transformer& W-A      &  OA(\%)   & mAcc(\%)      \\
			\hline
                \hline
			PointNet \cite{pointnet}        &\ding{55}& 32-32    & 89.2 & 86.0     \\
			\hline
			PointNet++ \cite{Qi2017PointNetDH} &\ding{55}& 32-32 & 90.7 &  -          \\
			\hline
			DGCNN \cite{Wang2019DynamicGC}    &\ding{55} & 32-32 & 92.9 & 90.0       \\
			\hline
			PCT \cite{guo2021pct}             &\ding{52} & 32-32 & \textbf{93.2} & \textbf{90.2}    \\
			\hline
            \hline
      	BiPointNet \cite{qin2020bipointnet} &\ding{55}&  1-1  & 86.4 &  -        \\
			\hline
			\bf{FBPT(Ours)}                    &\ding{52} &  1-1  & \textbf{90.9} & \textbf{87.8}  \\ 
			\hline
			\hline
		\end{tabular}
	\end{table}

    % Baseline network
	\begin{table*}[t]
    	\caption{Comparison results on the average recall at top 1\% and top 1 of different baseline networks trained on Ox. and tested on Ox., U.S., R.A. and B.D., respectively.} 
		\label{tab:baseline-result}
		\centering
        %\rowcolors{1-3}{gray!50}
		\rowcolors{4}{gray!50}{gray!20}
		\begin{tabular}{c|c|c|c|c|c|c|c|c|c|c}  %确定表格竖行格式   设置列宽度：将c换为p{ cm}
			\hline       %加横线
			\hline
			%	\rowcolor{red}    %加入颜色
			%		\diagbox{}{} 
			\rowcolor{red!30}
            && & \multicolumn{4}{c|}{Ave recall @ 1\%} & \multicolumn{4}{c}{Ave recall @ 1}\\    
			%		\hline   
			\cline{4-11} \rowcolor{red!30}
			\multirow{-2}{15em}{\centering Methods}&\multirow{-2}{5em}{\centering Transformer}&\multirow{-2}{4em}{\centering Binary Operation}& Ox. & U.S. & R.A. & B.D. & Ox. & U.S. & R.A. & B.D. \\
			%		& \multicolumn{2}{c|}{Country List} \\     
			\hline
            \hline
			PN\_VLAD \cite{Uy2018PointNetVLADDP} &\ding{55} &\ding{55} &80.33&72.63&60.27&65.30&62.76&65.96&55.31&58.87\\
			\hline
			%	\rowcolor[gray]{0.8}
			PCAN \cite{Zhang2019PCAN3A} &\ding{55}&\ding{55}&83.81&79.05&71.18&66.82&69.05&62.50&57.00&58.14\\
			\hline
			DH3D-4096 \cite{du2020dh3d} &\ding{55}&\ding{55}&84.26&-&-&-&73.28&-&-&-\\
			\hline
			DAGC \cite{DAGC}&\ding{55}&\ding{55}&87.49&83.49&75.68&71.21&73.34&-&-&-\\
			\hline
			LPD-Net* \cite{Liu-LPD-Net}&\ding{55}& \ding{55}&91.61&86.02&78.85&75.36&82.41&77.25&65.66&69.51 \\
   		%\hline

			\hline			
			HiTPR \cite{hou2022hitpr}&\ding{52}&\ding{55} &93.71&90.21&87.16&79.79&86.63&80.86&78.16&74.26 \\ 
   		\hline
			EPC-Net \cite{Hui2021Efficient3P}&\ding{55}&\ding{55} &94.74&96.52&88.58&84.92&86.23&-&-&- \\
			\hline
            \hline
   		PCT \cite{guo2021pct} &\ding{52}&\ding{55}                     
                &93.26&89.72&87.11&78.81&85.98&80.43&77.56&74.01 \\ 
			\hline
            FBPT (ours) &\ding{52}&\ding{52} &91.02&86.33&82.76&76.42&82.87&78.31&73.01&71.95 \\ 
            \hline
            \hline
        \end{tabular}
	\end{table*}
    \noindent\textbf{Place recognition implementation details:} The network structure for the place recognition follows the structure used for classification except for the classifier header. In our experiment, a simple max-pooling layer for aggregating a globally consistent descriptor after the last linear layers for class labels generation are removed.
    With respect to the loss function, the metric learning loss is required for the place recognition task. Specifically, the loss function we used is a role for minimizing the relative distance between descriptors extracted from point clouds sampled at the revisited place (defined as positive samples) and maximizing the relative distance between descriptors extracted from point clouds sampled at different places (defined as negative samples) in the metric space. The lazy quadruplet loss proposed in \cite{Uy2018PointNetVLADDP} is used in our work.
    
    The dimension of the global descriptor produced by the last pooling layer is set to 256. We feed the final descriptors into the lazy quadruplet loss function and set the hyper-parameters $\gamma$ and $\theta$ in the loss function to 0.5 and 0.2, respectively. We wrap 1 anchor sample, 2 positive samples, and 8 negative samples together as a mini-batch and synchronously feed them into an NVIDIA A6000 GPU card. The optimizer Adam \cite{kingma2014adam} with an initial learning rate of $5\times10^{-5}$ is used to train our network $16+4$ epochs (two stages) iteratively on the Pytorch platform. The learning rate decays to $1\times10^{-5}$ eventually.

	\subsection{Quantitative Results}
	\label{subsec:main_result}
    \noindent\textbf{Classification results: }Experimental results on point cloud classification are shown in Tab. \ref{tab:result_tmp}. The bit number of the weights and activations is marked in the W-A column. In this table, PCT means full-precision point cloud transformer which is the counterpart of our binarized model. We compare our binary point cloud transformer with three kinds of methods, the first one is the full-precision method without transformer mechanism, such as PointNet\cite{pointnet}, DGCNN\cite{Wang2019DynamicGC}, \emph{etc.}, the second one is the full-precision transformer method (PCT\cite{guo2021pct}), and the last is the binary model (BiPointNet\cite{qin2020bipointnet}) binarized from PointNet \cite{pointnet}. This experiment demonstrates the proposed FBPT model achieves only a 2.3\% performance drop compared to the full-precision PCT model and even outperforms some full-precision models.
    
    \noindent\textbf{Place recognition results: }The results of the compared methods are shown in Tab. \ref{tab:baseline-result}. It demonstrates our proposed FBPT model can achieve competitive results with the existing full-precision models with smaller memory space and lighter calculation load. 
    It should be noted that the LPD-Net* indicates the handcrafted features are removed from the original LPD-Net \cite{Liu-LPD-Net} for a fair comparison. We remove the handcrafted features based on the fact that most previous works in LiDAR-based place recognition do not concatenate additional handcrafted features as their network input. 
	
    \subsection{Computational Cost Analysis}
    \label{subsec:com_cost}
\begin{figure}[t]
		\centering
		\subfigure[model size (87.2\%)]{
			\begin{minipage}[t]{0.22\textwidth}
				\centering
				\includegraphics[width=1\textwidth]{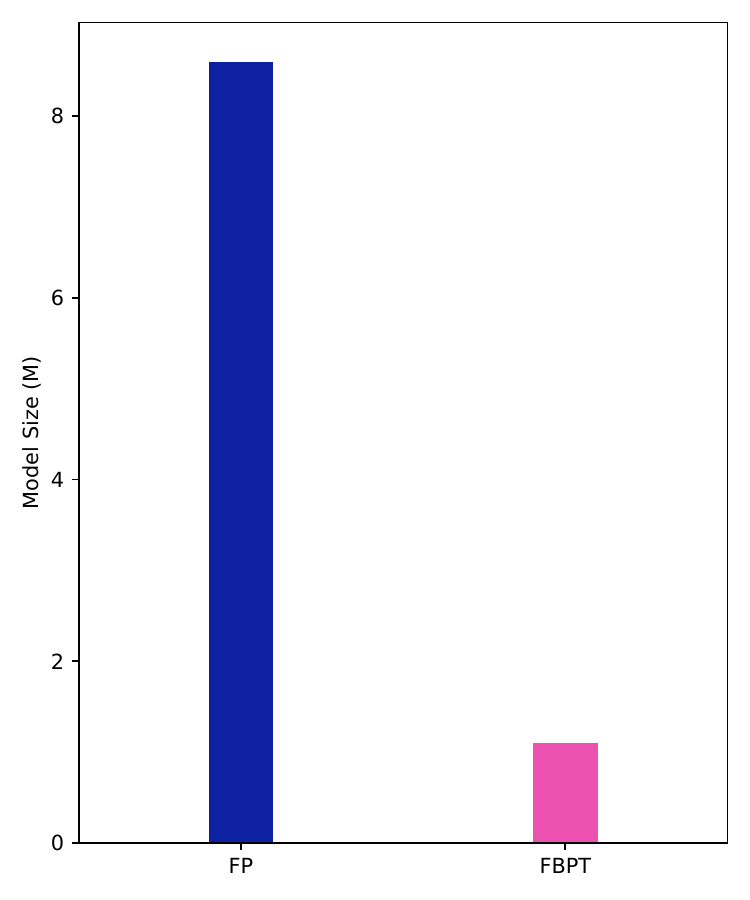}
				%\caption{fig1}
			\end{minipage}%
		}%
		\subfigure[FLOPs (80.2\%)]{
			\begin{minipage}[t]{0.22\textwidth}
				\centering
				\includegraphics[width=1\textwidth]{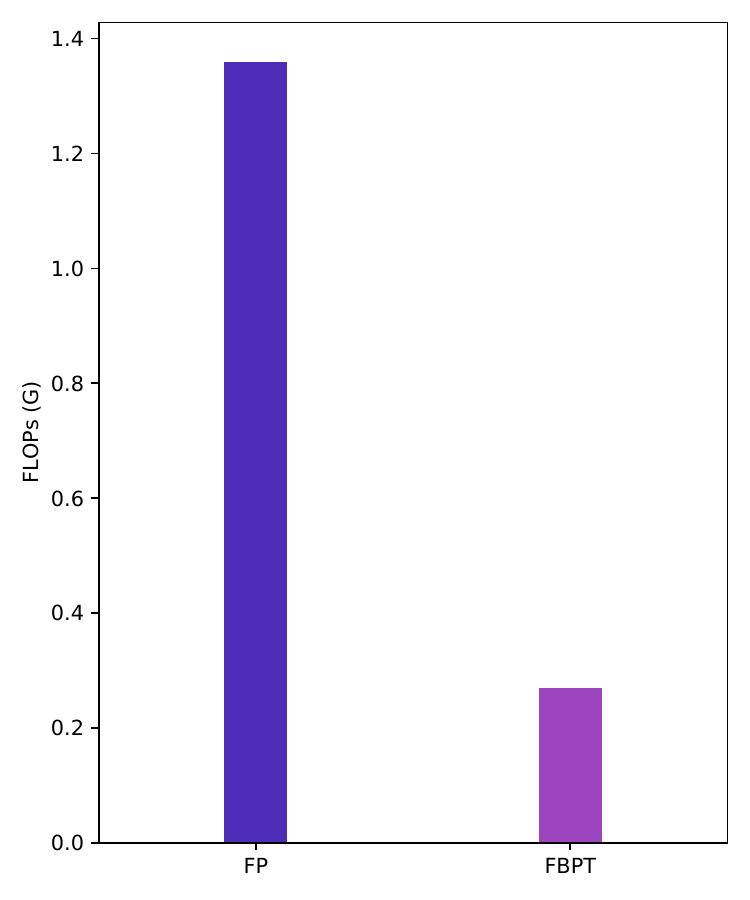}
				%\caption{fig2}
			\end{minipage}%
		}%
		\centering
		\caption{Calculation resource consumption comparison.}
		\label{fig:resource}
	\end{figure}

        We analyze the consumption of computing resources from two aspects: the model size and FLOPs (floating point operations). After binarization, the size of the same model structure reduces 87.2\% from 8.6M to 1.1M, while the FLOPs reduce 80.2\% from 1.36G to 0.27G shown in Fig.~\ref{fig:resource}. Compared with the full-precision point cloud transformer, the proposed binary one is more suitable to be deployed on real-world robots or mobile devices.

    \subsection{Contribution to Generalization}
    \label{subsec:generalization}
    
    In the experimental analysis, we observed that the full-precision vanilla transformer had a limited contribution to the model. Therefore, we investigated the impact of removing the binarized transformer module (referred to as $FBPT\_B$) from the FBPT model to assess its significance (referred to as $FBPT\_WO$). We discovered that the binarized transformer module led to improvements in result accuracy shown in ~\cref{tab:generation}. Moreover, more importantly, results on three small non-training datasets indicated that $FBPT\_B$ was able to mitigate overfitting and enhance the model's generalization capability
	
 \begin{table}[t]
    	\caption{Contribution to generalization.} 
		\label{tab:generation}
		\centering
        %\rowcolors{1-3}{gray!50}
		\rowcolors{4}{gray!50}{gray!20}
		\begin{tabular}{c|c|c|c|c}  
			\hline       
			\hline

			%		\diagbox{}{} 
			\rowcolor{red!30}
            & \multicolumn{4}{c|}{Ave recall @ 1\%} \\   
			%		\hline   
			\cline{2-5} \rowcolor{red!30}
			\multirow{-2}{6em}{\centering Methods}& Ox. & U.S. & R.A. & B.D.  \\
			%		& \multicolumn{2}{c|}{Country List} \\     
			\hline
            \hline
		    FBPT\_WO  &89.84&82.16&76.77&70.41\\
			\hline
            FBPT\_B &91.02&86.33&82.76&76.42 \\ 
            \hline
            \hline
        \end{tabular}
	\end{table}	

	%===============================================================================
	
    \section{Conclusion}
    \label{sec:conclusion}
    
        This paper presents a novel Fully Binary Point Cloud Transformer model. To address the performance degradation issue caused by the use of binary point cloud Transformer modules. We propose fine-grained binarization, dynamic-static hybridization, and a hierarchical training scheme. With these improvements, we obtained excellent experiment results on point classification and place recognition tasks.
        Although there are currently limited devices that can support binary operations efficiently, research in this direction still holds scientific value and provides valuable guidance. Furthermore, the quantization experimental results further demonstrate the great potential and advantages of our model in real-world point cloud tasks, particularly on resource-constrained devices.

	%===============================================================================

    %\noindent\textbf{Acknowledgements}: xxxx

    % \clearpage 

    \bibliographystyle{IEEEtran}
    \bibliography{example}  % .bib
	
\end{document}